# OAIR: Object-Aware Image Retargeting Using PSO and Aesthetic Quality Assessment

Mohammad Reza Naderi[1], Mohammad Hossein Givkashi[*1], Nader Karimi[1], Shahram Shirani[2], Shadrokh Samavi[1,2]

[1] Department of Electrical and Computer Engineering, Isfahan University of Technology, Isfahan, Iran
[2] Department of Electrical and Computer Engineering, McMaster University, Hamilton, Canada

*Abstract*— Image retargeting aims at altering images' size while preserving important content and minimizing noticeable distortions. However, previous image retargeting methods create outputs that suffer from artifacts and distortions. Besides, most previous works attempt to retarget the background and foreground of the input image simultaneously. Simultaneous resizing of the foreground and background causes changes in the aspect ratios of the objects. The change in the aspect ratio is specifically not desirable for human objects. We propose a retargeting method that overcomes these problems. The proposed approach consists of the following steps. Firstly, an inpainting method uses the input image and the binary mask of foreground objects to produce a background image without any foreground objects. Secondly, the seam carving method resizes the background image to the target size. Then, a super-resolution method increases the input image quality, and we then extract the foreground objects. Finally, the retargeted background and the extracted super-resoluted objects are fed into a particle swarm optimization algorithm (PSO). The PSO algorithm uses aesthetic quality assessment as its objective function to identify the best location and size for the objects to be placed in the background. We used image quality assessment and aesthetic quality assessment measures to show our superior results compared to popular image retargeting techniques.

*Index Terms*— Aesthetic quality assessment, image retargeting, Particle swarm optimization.

## 1. Introduction

Image retargeting is aesthetically one of the essential tasks in computer vision. This task is about making images capable of being displayed on different screen sizes by considering high-level features of the image. In other words, image retargeting aims to resize an image effectively and preserves important objects without distorting the image's textures, colors, and boundaries. Multiple efforts have been made to handle the image retargeting problem during the last decade. One of the most noticeable traditional methods is the seam-carving algorithm [1]. It extracts the image energy map and considers pixels with a close amount of energy between rows as seams, then by adding or removing the low energy seams, attempting to change the image size [1].

Deep learning models with convolutional neural networks achieved impressive results in computer vision tasks, and image retargeting is no exception. The most crucial bottleneck to training deep models to retarget an image is that these models require thousands of paired images to learn the desired image translation effectively. Some researchers attempted to create paired datasets to train deep models in a weakly supervised manner [2]. Weakly-supervised learning was used since we cannot define the ground truth for a given image in multiple sizes because it is quite a matter of taste. The desired image for each person or task could be completely different. Therefore, achieving a paired dataset is the main obstacle to developing deep learning models for image retargeting. Some one-shot learning approaches are proposed to solve the issue of not having paired images [3, 11]. But one-shot learning does not work efficiently because the model must be trained for each image separately [3].

Furthermore, the total time to resize an image is essential [4]. For example, when we want to transfer an image from a smartphone to a personal computer, we do not have the required hardware to train a model. Also, one cannot tolerate a long delay in opening up an image on their personal computer. However, some other works used deep learning in a weakly or self-supervised manner to learn the task of image retargeting by using conceptual losses [5, 6]. These proposed methods are time-consuming because they require to be trained for each image. Also, the quality of generated images is not always satisfactory. Based on the shortcomings of the previous learning methods, the need for a comprehensive, fast and accurate method in this field is felt, which could be applied to natural images with various foreground objects. In this work, we propose an image retargeting method. We aim to perform image retargeting in conjunction with other tasks such as image inpainting, seam carving, super-resolution, particle swarm optimization, and aesthetic quality assessment. Using these image processing tasks, we focus on objects in the image and process the background and foreground of the image to create an aesthetically acceptable retargeted image with the desired size.

The structure of this paper is as follows: In section 2, we will present some previous retargeting methods. Then, in section 3, also we will introduce some machine-vision tools that we will use in the rest of the paper. Then, section 4 offers the proposed method. Finally, experimental results are detailed in section 5, and concluding remarks are presented in section 6.

## 2. Related Works

In this section, we will look at some existing retargeting methods. Like most computer vision tasks, image retargeting research could also be classified as rule-based and deep learning-based methods using algorithmic and learning

* *The first two authors contributed equally

approaches. We will discuss the proposed works in each category below.

*A. Rule-Based Retargeting Methods*

Traditional image retargeting methods were based on image processing techniques. Examples of retargeting methods include seam carving for content-aware image resizing [1] and multi-operator media retargeting [7]. The method presented in [7] extracts the image energy map and considers pixels with similar energies between rows as a seam. Then, it adds or removes the low-energy seams to change the size of the image. The main difference between these two methods is that the first one considers continuous seams, and the second one shows that considering discontinuous seams could yield better results. In [8], the authors proposed a method for increasing the speed of the seam carving algorithm. They used multi-GPU to speed up the proposed algorithm to reduce computation time.

*B. Learning-Based Retargeting Methods*

There are multiple limitations to training deep learning models for image retargeting tasks (which we will discuss). However, various works have attempted to use different learning models. We will discuss some of the learning models below.

Donghyeon Cho [5] proposed a weakly- and self-supervised deep learning model which used an encoder-decoder architecture trained on PASCAL VOC 2007 dataset [9]. The encoder part of the network extracts high-level information from images, and the decoder part uses that information to generate an attention map. Then, the model creates desired output based on the regions that the decoder considers essential components. Weimin Tan [6] proposed a deep-cyclic framework for image retargeting. Their architecture contains a cycle that has been used to reconstruct the input from the output. The input reconstruction helps the model save as much important information as possible while producing the output. They also used attention modules to extract an attention map from the input image. Spatial and channel attention can help the model focus on essential features from the input image and generate better output results.

A generative adversarial network (GAN) is an effective model for unsupervised learning in computer vision [10]. Shocher [3] used a generative adversarial network for image retargeting. They proposed a one-shot learning GAN which must be trained for each image separately. The model learns the internal distribution of an image and, when trained once on an image, could retarget that image to different sizes. Due to learning internal image distribution by using techniques such as cropping and adding noise to the image, their model does not entirely understand the whole scene. The model duplicates every object when targeting the image to a bigger size. Hence, in most cases, the objects are destructed in smaller retargeted images. They also showed a failure example when the model wants to retarget a human object and undesirably duplicates the person in the output image. Besides, the training procedure is time-consuming, between 1.5 to 4 hours on a single Nvidia Tesla V100 GPU for a single image. It seems that the time that their model takes to train on a single image is not empirical in real-world applications. In [11], an adversarial learning approach is used for tasks such as image super-resolution, denoising super-resolution, and image retargeting. In the image retargeting part of their works, their model learns the internal distribution of an image, similar to [3], to retarget the input image to different sizes. Their retargeting approach has the same problems as [3] as was discussed. Ya Zhou [12] proposed a method based on reinforcement learning for image retargeting. They defined the retargeting task as a sequence of operations that must be executed on the image, such as seam carving, scaling, and cropping. These three operators are considered actions, and the reinforcement learning model was rewarded using semantic and aesthetic measurements. Their model attempts to learn a sequence of actions greedily and, in some cases, results in aesthetically acceptable outputs.

The inference time, in real-world problems, is one of the critical factors of the image retargeting methods. In [13], Smith proposed a model that has been used in mobile devices and desktops. Using a content-aware cropping algorithm, they used image semantic segmentation and saliency detection to extract an importance map from the image to generate the output. As a result, retargeting execution times are reported on desktop and mobile devices, showing their advantage over time-consuming methods. In [2], two problems for image retargeting are mentioned. First, there are no comprehensive datasets with ground truth for supervised learning. The second is that almost all methods have severe limitations that prevent them from generating acceptable results in every image retargeting scenario. The authors used Image Retargeting Quality Assessment (IRQA) algorithm to evaluate outputs that multiple models generated, and they selected the best one as ground truth. Then, they attempt to train a GAN model on this dataset in a weakly supervised manner. Therefore, the proposed methods are not efficient for real-world application, and in some cases, they do not generate acceptable-quality images. To overcome and solve these problems, we proposed an image retargeting framework that focuses on the objects in the images and generates high-quality outputs.

In [14], the authors have proposed a framework for single-image generative tasks. They generate output with a single image faster than single-image GAN-based methods. Their framework has been used in tasks such as image retargeting, diverse image generation, and structural analogies. They used the patch nearest neighbor approach and introduced a casting GPNN, a generative patch-based algorithm, which has been used in different tasks. Their study demonstrated that casting of classical methods results in novel and high-quality images that are faster than those produced using single image GANs. An algorithm for minimizing the distance between patch distributions in an image has been developed in [15] based on the recently developed Sliced Wasserstein Distance. Their GPDM algorithm uses patches from the input image to create output. They used the algorithm in different tasks like, style transfer, image editing, image retargeting and etc.

## 3. Applicable Vision Tools

This section briefly discusses the basic tools used in our work. These concepts include image inpainting, seam carving, super-resolution, particle swarm optimization, and aesthetic quality assessment. We also look at pioneer works in each field we used in our proposed method.

*A. Image Inpainting*

Image inpainting is the process of filling holes or empty regions in images or removing specific objects from an image [16-19]. It is important for a model that fills empty regions in an image to produce a good result by matching the features of the generated region and input part. A method for image inpainting focused on ensuring consistency in color has been proposed in [17]. This method generates reliable outputs with no artifacts. Stable colors in various parts of the image can help achieve a realistic appearance. One of the most difficult tasks in image inpainting is face adaptation. Face adaptation involves changing the face from one image to another. Face inpainting is complex because the face must be consistent with the picture's light, position, hairstyle, and other elements. In [18], they used some pre-trained networks to identify faces and textures from images. This method helps the model to inpaint face images more effectively than previous methods. A general view of an image could help the model fill a particular image region more effectively. In [16], the authors used Fast Fourier Convolution in their proposed method for image inpainting. With the help of Fourier transformation, they increased the global view of their model to generate better output results. Their model is capable of filling different blanked shapes on the image. Since they did not apply any restriction on the shape that the model could fill, it generalizes well to unseen images.

*B. Image Super-Resolution*

Image super-resolution involves increasing the resolution and quality of an image [20-23]. The real-world images contain noise, have low resolution, and suffer from compression artifacts. The model must be trained with similar pictures to increase the quality of these images by using a learning approach. In [23], the authors used training images after applying noise, downsampling, and compression to images. The proposed model is based on a generative adversarial network. In [22] also authors proposed a deep learning model for real-world super-resolution. Their model could handle noisy or low-quality images that exist in real-world applications. In the past two years, transformers almost dominated this task with their power to generate high-quality outputs. The Swin transformer is one of the transformer-based architectures that has been used in multiple tasks such as image segmentation and image inpainting. SwinIR [20] uses Swin transformer [21] for image restoration, including image super-resolution, image denoising, and JPEG compression. The SwinIR method has shallow feature extraction, deep feature extraction, and high-quality image reconstruction modules. The SwinIR method has achieved interesting results in performing the mentioned tasks.

*C. Aesthetic Quality Assessment*

Aesthetic quality assessment schemes evaluate the quality of an image based on beauty from the perspective of the human eye [24-25]. There are multiple research works in this category. However, we chose the one that could handle input with different sizes (without resizing input). Therefore, using a method that takes any size input would help us evaluate the quality of any retargeted image. In [25], the authors proposed an aesthetic quality assessment method that can assess input images of any size and thus is appropriate to evaluate the quality of retargeted images.

*D. Particle Swarm Optimization (PSO)*

Searching a space to find the most fitted results, considering a non-differentiable fitness function, is typically done using heuristic and meta-heuristic search algorithms [26-28]. Particle swarm optimization (PSO) [28] is one of the most successful algorithms in this field that attempts to simulate birds' flocking to optimize the fitness function in a given space.

## 4. Proposed Method

Image retargeting aims to resize an image and preserve the critical elements of the scene (foreground) without any visible distortion in the resized image. Seam carving is one of the powerful methods to retarget an image. However, the main drawback of the seam carving approach is that when it attempts to add (remove) a seam to (from) the image, it shifts the pixels. As a result, seam carving causes severe distortions in the borders of the objects. Examples of seam carving distortions are shown in Fig. 1(a). Distortions are more visible when the desired target size is very much different from the image's size. However, for an image without many edges, in that case, the seam carving could easily resize the image to the desired size without distortion, as shown in Fig. 1(b). This observation led us to propose a method that processes the foreground and background of the image separately. After processing, we merge the foreground and background to get the highest fitness from the aesthetic point of view. Fig. 2 shows an overview of our proposed method. As we can see, we have several stages to generate our desired retargeted output. First, we will discuss the processes done on the foreground and the background in subsections IV.A and IV.B, respectively. Then, in subsection IV.C, the steps used to produce output will be discussed. To summarize our method, first, we extract the foreground and background of the image. By using the inpainting model, the object's location in the background is filled. Following that, the seam carving method is utilized to generate a background of the desired size. Then, the visual quality of the foreground is increased by employing a super-resolution algorithm. Next, particle swarm optimization and aesthetic quality assessment are used to determine the best location and size of the foreground image. Finally, we merge the foreground and background to generate a high-quality, aesthetically pleasing output.

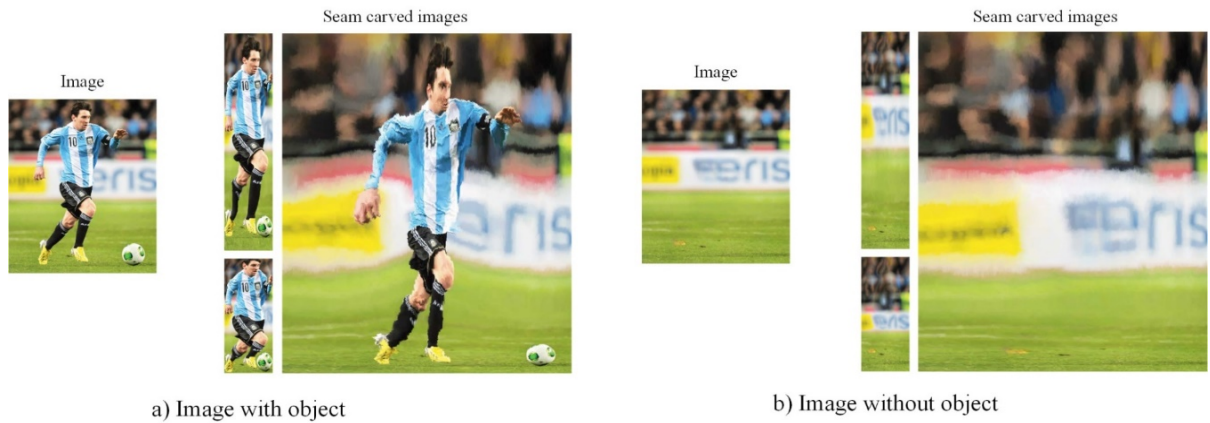

**Fig. 1.** Comparing seam carving performance to retarget an image to different sizes. a) An image with a foreground object, b) an image without an object. The resizing factors are randomly selected between 0.33 and 2 and multiplied by the input image height and width independently

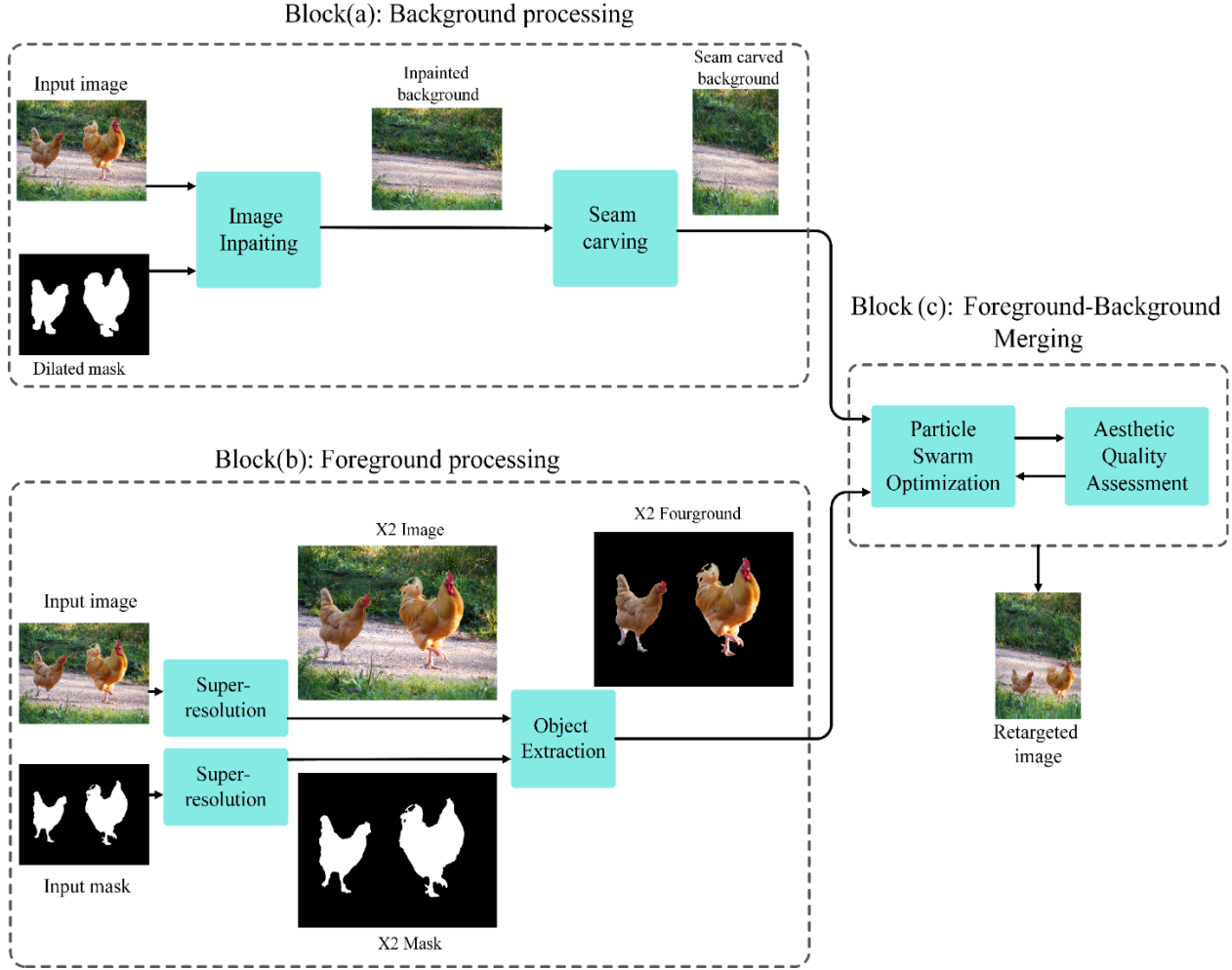

**Fig. 2.** Block diagram of the proposed method. Block(a) foreground processing, block (b) background processing, and block(c) performs foreground-background merging.

*A. Background Processing*

As we can see in block (a) of Fig. 2, firstly, we feed the input image and a dilated binary mask of the foreground to the image inpainting block. In this stage, we used a pre-trained image inpainting model called big-LaMa [16]. Big-LaMa uses Fast Fourier Convolution layers to achieve a global understanding of the image. As a result, it generates high-quality inpainted images. Big-LaMa was trained on an extensive natural image dataset. Thus, it can generalize well almost every image inpainting problem. Big-LaMa applies the mask to the image and then performs inpainting of the generated empty regions using the surrounding pixels. In other words, it removes the objects from the image and fills out the missing regions. The foreground masks in the segmented datasets are generally not precise around the boundaries of the objects. Hence, the mask

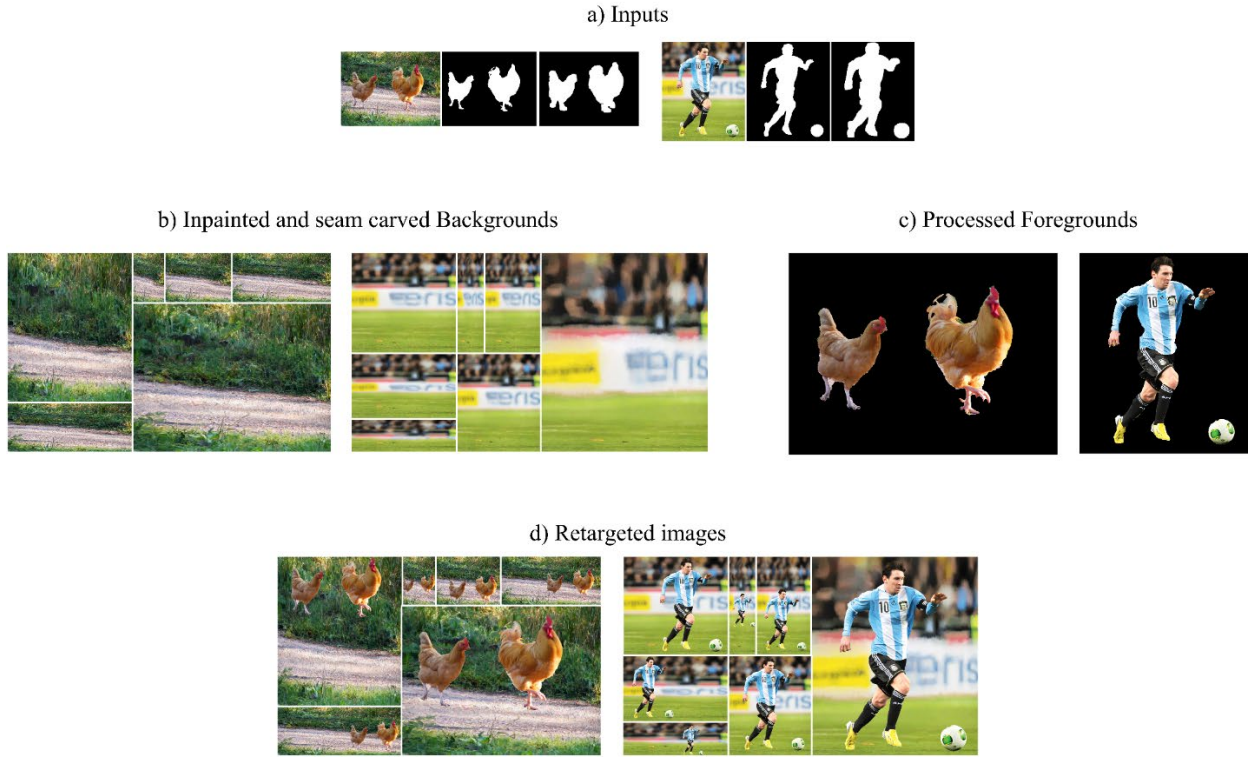

**Fig. 3.** The procedure that we proceed to retarget Inputs. a) Inputs b) Omitting foreground and Inpainting and seam carving backgrounds. c) Performing super-resolution on input images, then extracting the foreground. d) Retargeting results by merging foreground and background using PSO and aesthetic quality assessment

is dilated by performing a morphological dilation operation to prevent the remaining object pixels from downgrading the image inpainting quality. Afterward, the inpainted background is fed as input to the seam carving algorithm [1]. This algorithm extracts seams based on the image's content and uses these seams to resize the background image to the desired output size (the results for this stage are shown in Fig. 3(b).

*B. Foreground Processing*

We use a super-resolution model [20] to increase the quality and size of the input image and the binary mask of the foreground. We then apply the super-resolued mask to the super-resolued image to extract a super-resolued high-quality foreground (the results for this stage are shown in Fig. 3(c). The main idea behind this super-resolution stage is that in some cases, we want retargeting algorithm to resize the image to a greater size than the original image, so we need the foreground of high quality for those greater sizes. Therefore, by using super-resolution, we now have a high-quality foreground (four times greater than the size of the original image), which is suitable for the retargeting range. These processes are shown in block (b) of Fig 2.

*C. Foreground - Background Merging*

The critical part of this work is to merge the foreground and background. This is shown in block (c) of Fig. 2. Finding a suitable size for the foreground and the best position to place the foreground on the background is done using a search algorithm and an aesthetic quality assessment model. The aesthetic quality assessment is a useful way to measure the attractiveness of images. Therefore, we use the aesthetic quality assessment as the objective function of our search algorithm.

The aesthetic assessment model gets one image as input and generates a score in the output that shows the aesthetic quality of the image. However, the search space (size and position in the foreground) is enormous. Therefore, we use Particle Swarm Optimization (PSO) [28] to reduce search space and get results in an efficient time. As our search algorithm is an iterative procedure, first, the PSO chooses a position and size for the foreground. Then we merge the foreground with the background using the selected position and size. Then we find an aesthetic score for the image that the aesthetic quality of an image is assessed based on its beauty. This procedure is continued until the PSO algorithm converges to the most fitting answer to the problem. Examples of the results for this stage are shown in Fig. 3(d).

Fig. 4 shows the size and position that PSO examined until the convergence. It also shows some high-quality merged foregrounds and backgrounds in different sizes found during the search procedure.

A pseudo-code for our proposed method is also presented in Algorithm 1, which provides more insight into the work done to retarget an image. Our algorithm takes the image, mask, dilated mask, and the desired target size. Then, the explained steps in the background and foreground sections generate the desired output.

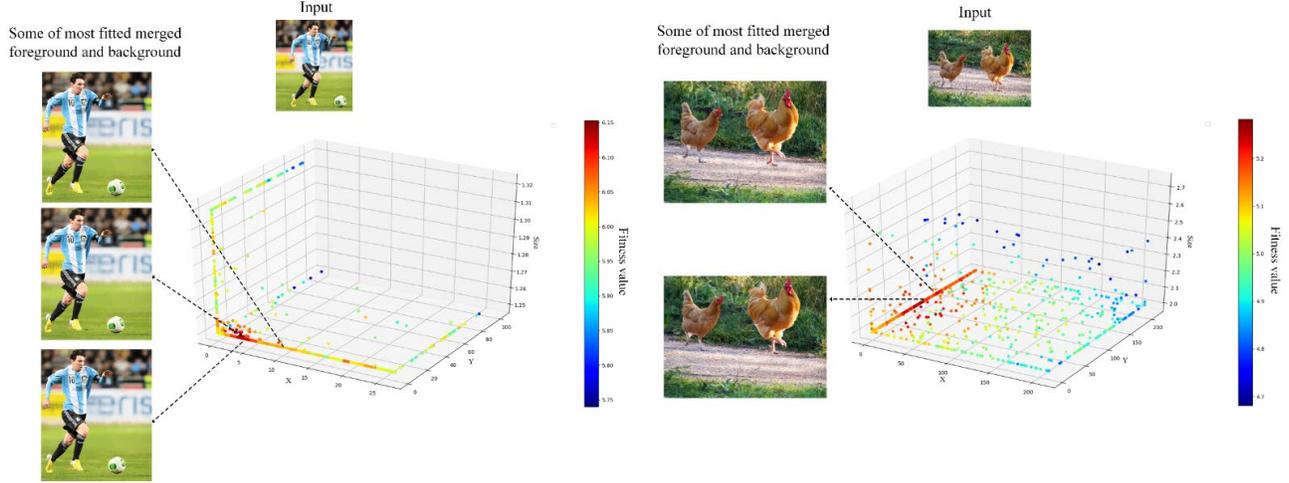

**Fig. 4.** Some of the most fitted merged images of the foreground (FG) with the background (BG). Points that PSO selected during maximizing aesthetic quality assessment in the X (vertical FG position in BG), Y (horizontal FG position in BG), and Size (size of FG that is placed on BG) to twice Input image.

| **Algorithm 1**: Pseudo code of the proposed method. |
|---|
| Abbreviations: <br> AQA: Aesthetic Quality Assessment <br> D: Dilation <br> IP: Inpainting <br> SC: Seam Carving <br> SR: Super-Resolution <br> FBM: Foreground and Background Merging <br> *bg: image background* <br> *fg: foreground objects* <br> *x: vertical position of FG on BG* <br> *y: horizontal position of FG on BG* <br> *size: the size of FG* |
| Algorithm: <br> Input data: *image, mask, target_size* <br> *d_mask* = D(*mask*) <br> *ip_bg* = IP(*image, d_mask*) <br> *sc_bg* = SC(*ip_bg, target_size*) <br> *sr_fg* = SR(*image, mask*) <br> Randomize *x, y, size* <br> **While** PSO converges: <br>     *generated_image* = FBM(*sc_bg, sr_fg, x, y, size*) <br>     *fitness_values* = AQA(*generated_images*) <br>     [*x, y, size*] = PSO(*fitness_values*) <br>     **return:** FBM(*sc_bg, sr_fg, best_x, best_y, best_size*) |

## 5. Experimental Results

We created a new dataset containing images and their binary object segmentation masks to test our proposed method. The dataset also contains the outputs of our proposed method and other significant image retargeting methods with aspect ratios of 0.33, 0.66, 1.0, 1.25, and 2.0. The dataset is at https://github.com/givkashi/OAIR.

In this section, first, we qualitatively discuss the generated results using two different methods and our proposed methods. Then we will discuss the quantitative performance of each method using aesthetic and normal quality assessment pre-trained networks.

### A. Qualitative Results

Figures 5 and 6 show several visual examples of retargeting methods. The first row in each figure is the original image, and the following rows represent the seam carving [1], INGAN [3], and our proposed method's results. As expected, the seam carving method destroys the boundaries and margins around the objects. Therefore, the quality of output images decreases as we seam-carve images to greater or smaller sizes. The quality reduction is because the image pixels are shifted multiple times when inserting or removing a large number of seams. The movement of pixels affects the boundaries of objects as the shifting values could not be the same for all boundary pixels during the seam carving algorithm. Considering Fig. 5(b) and Fig. 6(b), we can see the drawbacks of the INGAN method. The outputs of this method are blurry because noise is used for augmentation, and as expected, the network has not completely understood the whole scene.

Thus, when the image gets larger, it attempts to duplicate the objects which are not desired, especially for human images. For smaller sizes, it almost destroyed the objects. We think this happened since their network does not perceive the essential and non-important parts of the image. Besides, as mentioned before, training a new model for each image does not seem efficient. In Fig. 5(c), we can see the output of our proposed method. Our method considered the importance of the object in the image. It does not change the aspect ratio of objects. The size of an object could change only linearly. Besides, the super-resolution of the object allows our method to save the object quality as the desired retargeting size increases. The used inpainting method also works well. Seam carving also does not have difficulties changing the size of objectless images. The results are quite remarkable for small or large retargeting sizes.

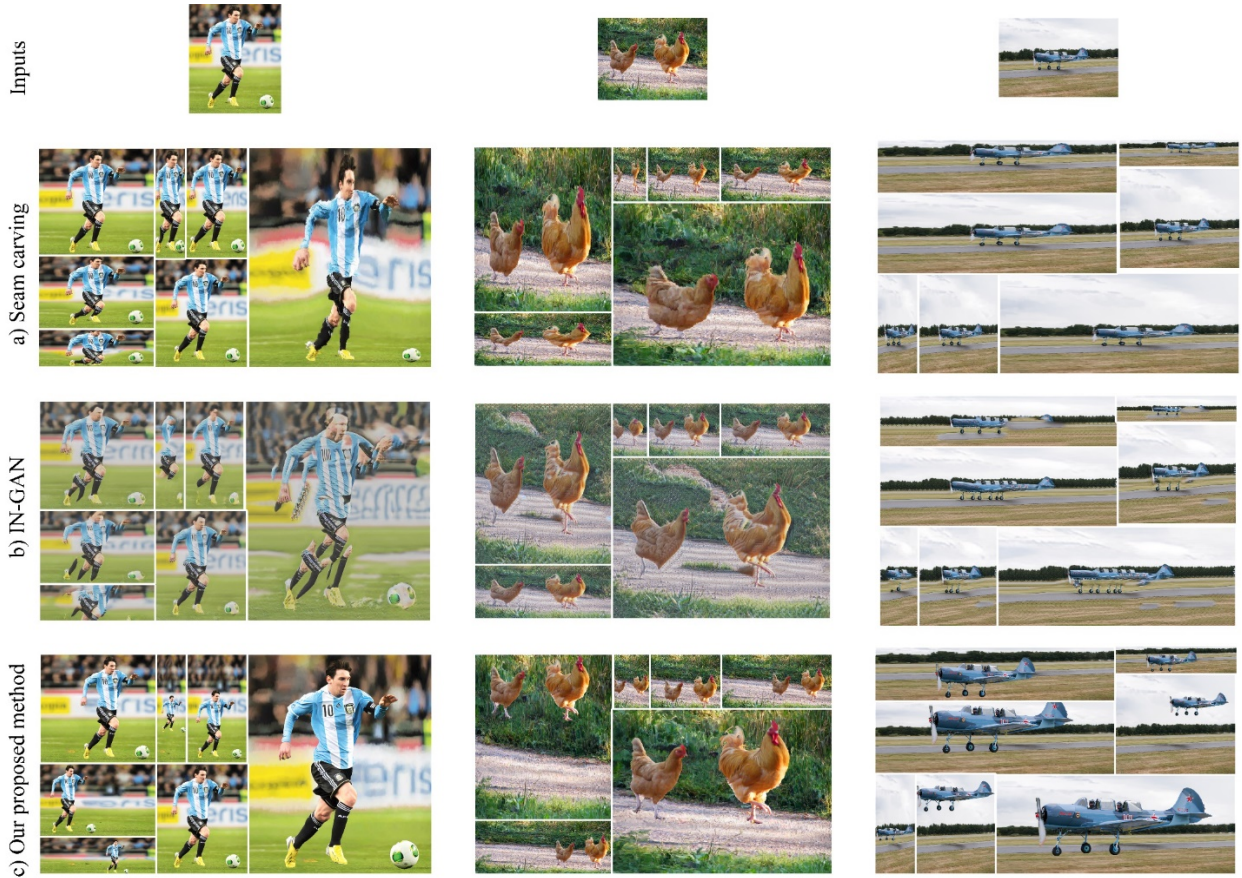

**Fig. 5.** Comparing results of our proposed method with seam carving [1] and InGAN [3] in multiple target sizes for images 1 to 3, which are adapted from [29-30]. The proposed retargeted image sizes are selected randomly between 0.33 and 2 multiplied by the real image height and width independently.

Figure 7 illustrates several examples of retargeted outputs. As shown in the first row, some examples of outputs are obtained using the GPNN [14] method for images 1 through 6. The second row shows some outputs from the GPDM [15] method, while the last row shows outputs from our proposed method. GPNN and GPDM use only input images to generate retargeted images. Both methods involve converting the input image into patches, which are then used to increase or decrease the size of the image. In the output images, we can see repeated objects and artifacts that result from this process. Therefore, processing background and foreground separately seems to be a better option in such cases. Thus, our method outperformed the mentioned methods in qualitative measurements.

*B. Quantitative Results*

We used aesthetic quality assessment [25] and a quality assessment pre-trained networks [31] to compare our method with resizing, seam-carving, and INGAN methods. TABLE I(a) shows the mean of aesthetics, and TABLE I(b) shows normal scores that each method achieved during the retargeting image to 25 different sizes. Better results are bolded. The generated image in real size for Images 1 and 2 are shown in Fig. 8. The aesthetic and normal scores of each algorithm are compared for all 25 different sizes in Fig. 9, which shows our method's superiority in most cases compared to the previously mentioned methods. TABLE II(a) shows the mean of aesthetics, and TABLE II(b) shows normal scores that were achieved for GPNN, GPDM, and our method for the retargeted images to 9 different sizes. Because the GPNN and GPDM methods could not handle small images, instead of 25, just 9 available retargeting sizes have been used in this table. The operation used by them is convolution, and it is not capable of running in small image sizes. The proposed retargeted image sizes are selected between 2, 1.25, and 1 multiplied by the real image height and width independently. We also show the meantime that each algorithm takes to generate the desired image in TABLE III. For INGAN, we reported the learning time for one image. Because each time that we want to retarget a picture, we need to train a new model. TABLE I, II, and III show that using our proposed method will result in a high-quality image in a reasonable time, which is more applicable than the learning base methods.

8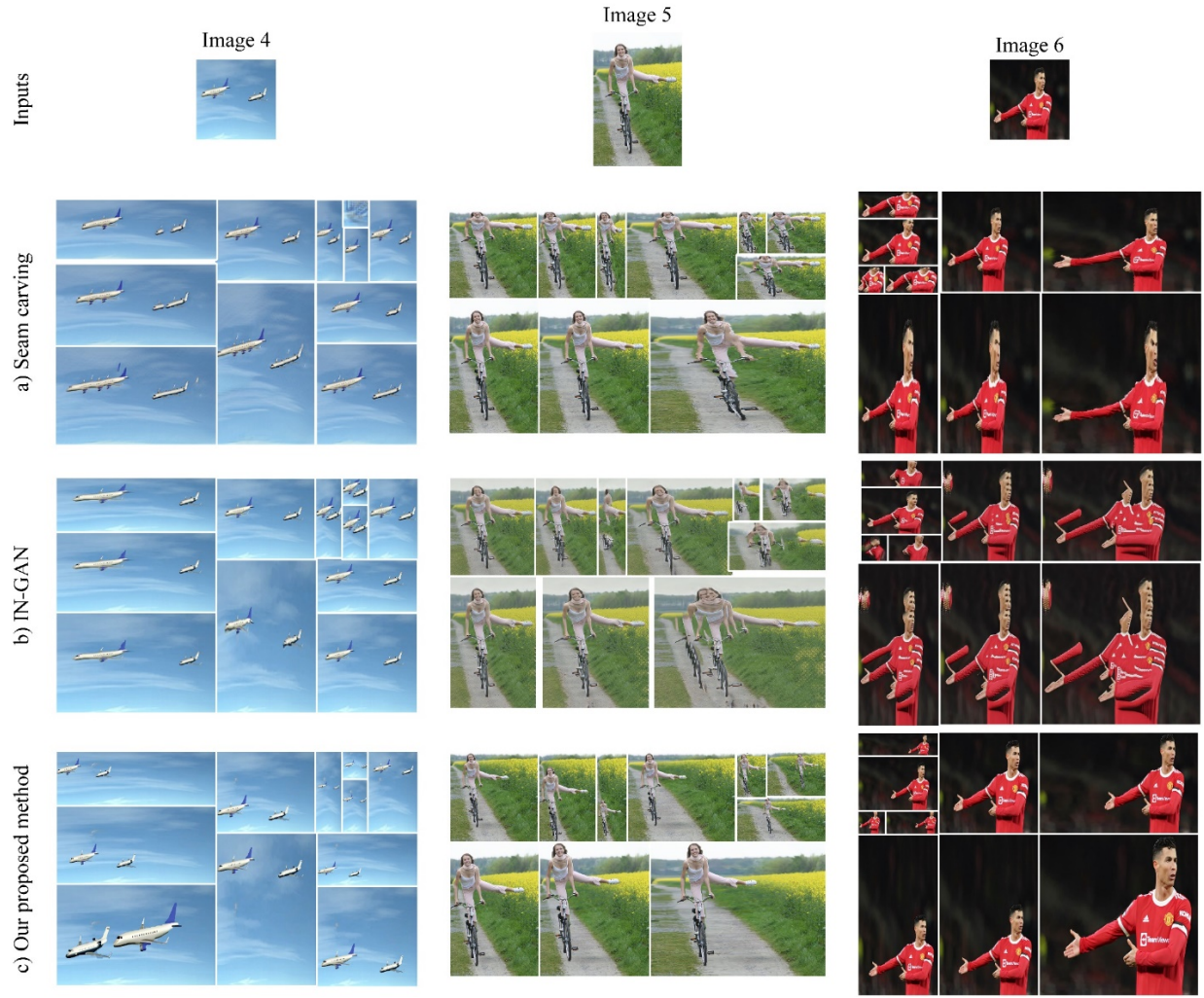

**Fig. 6.** Comparing results of our proposed method with seam carving [1] and INGAN [3] in multiple target sizes for Images 4 to 6, which are adapted from [29-30]. The proposed retargeted image sizes are selected randomly between the range (0.33, 2) multiplied by the real image height and width independently.

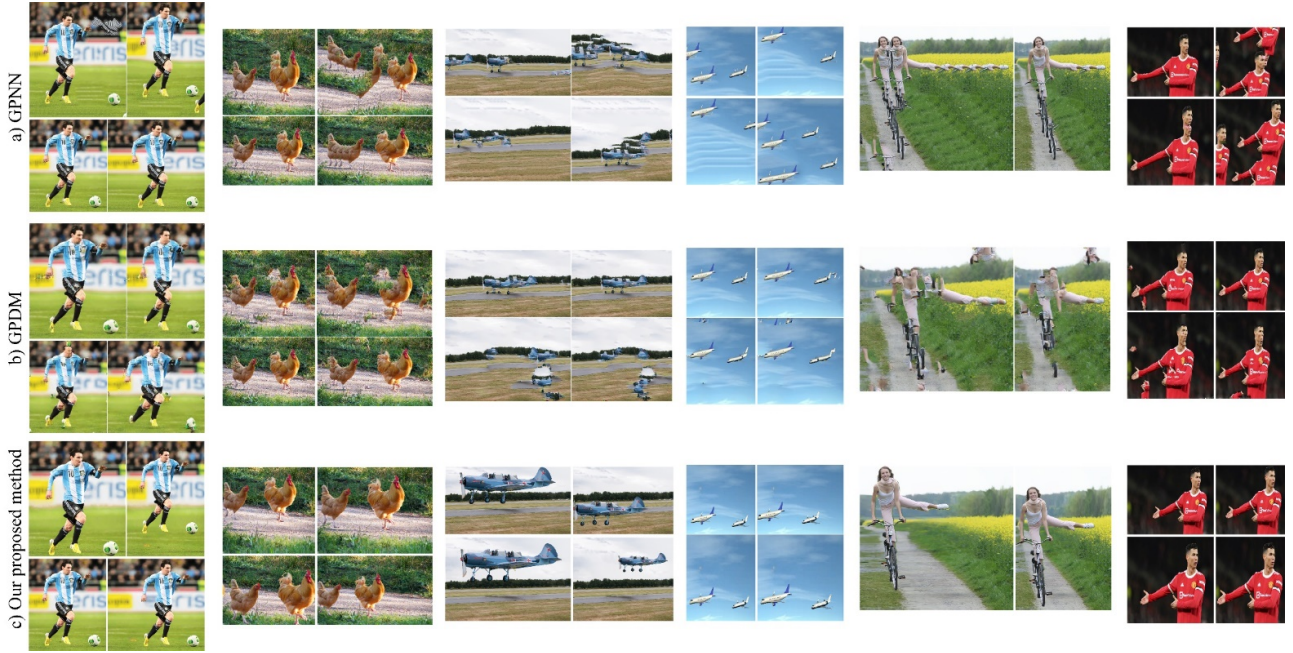

**Fig. 7**. Comparing results of our proposed method with GPNN [14] and GPDM [15] in multiple target sizes for Images 1 to 6, which are adapted from [29-30]. The proposed retargeted image sizes are selected between 2, 1.25 and 1 multiplied by the real image height and width independently

TABLE I
Comparing the performance of our proposed method with Resizing (bilinear), Seam carving [1] and INGAN [3] using the mean value of pre-trained aesthetic quality assessment and image quality assessment networks.

| Image | Resizing (bilinear) | Seam carving | INGAN | Ours |
|---|---|---|---|---|
| a) Aesthetic quality assessment using [25] | | | | |
| Image 1 | 4.24 | 4.62 | 4.09 | **5.27** |
| Image 2 | 3.05 | 3.40 | 3.31 | **4.05** |
| Image 3 | 4.01 | 4.13 | 4.12 | **4.81** |
| Image 4 | 3.01 | 2.98 | 2.97 | **4.11** |
| Image 5 | 4.21 | 4.65 | 4.15 | **5.01** |
| Image 6 | 3.60 | 3.62 | 3.33 | **4.12** |
| b) Image quality assessment using [31] | | | | |
| Image 1 | 52.26 | 60.19 | 57.89 | **60.90** |
| Image 2 | 52.29 | 63.86 | 62.12 | **68.66** |
| Image 3 | 49.49 | 60.99 | 51.34 | **67.21** |
| Image 4 | 47.84 | 51.29 | 51.21 | **64.50** |
| Image 5 | 57.29 | 64.22 | 65.19 | **66.65** |
| Image 6 | 51.43 | **57.68** | 53.17 | 56.42 |

TABLE II
Comparing the performance of our proposed method with GPNN [14] and GPDM [15] using the mean value of pre-trained aesthetic quality assessment and image quality assessment networks.

| Image | GPNN | GPDM | Ours |
|---|---|---|---|
| a) Aesthetic quality assessment using [25] | | | |
| Image 1 | **6.00** | 5.01 | **6.00** |
| Image 2 | 4.29 | 4.13 | **4.75** |
| Image 3 | 4.91 | 4.97 | **5.64** |
| Image 4 | 4.26 | 4.15 | **5.13** |
| Image 5 | 5.25 | 4.49 | **5.97** |
| Image 6 | 5.05 | 4.71 | **5.42** |
| b) Image quality assessment using [31] | | | |
| Image 1 | **62.54** | 34.92 | 61.71 |
| Image 2 | 64.01 | 58.09 | **64.14** |
| Image 3 | 56.85 | 53.41 | **62.82** |
| Image 4 | 50.84 | 45.33 | **63.49** |
| Image 5 | **66.33** | 34.13 | 64.55 |
| Image 6 | 60.22 | 54.05 | **61.86** |

TABLE III
Mean time to retarget an image to the desired size using different methods.

| Method | Resizing (bilinear) | Seam carving | INGAN | GPNN | GPDM | Ours |
|---|---|---|---|---|---|---|
| Time | 0.014 sec | 5.50 min | 7 hr | 4 min | 3 min | 7.47 min |
9

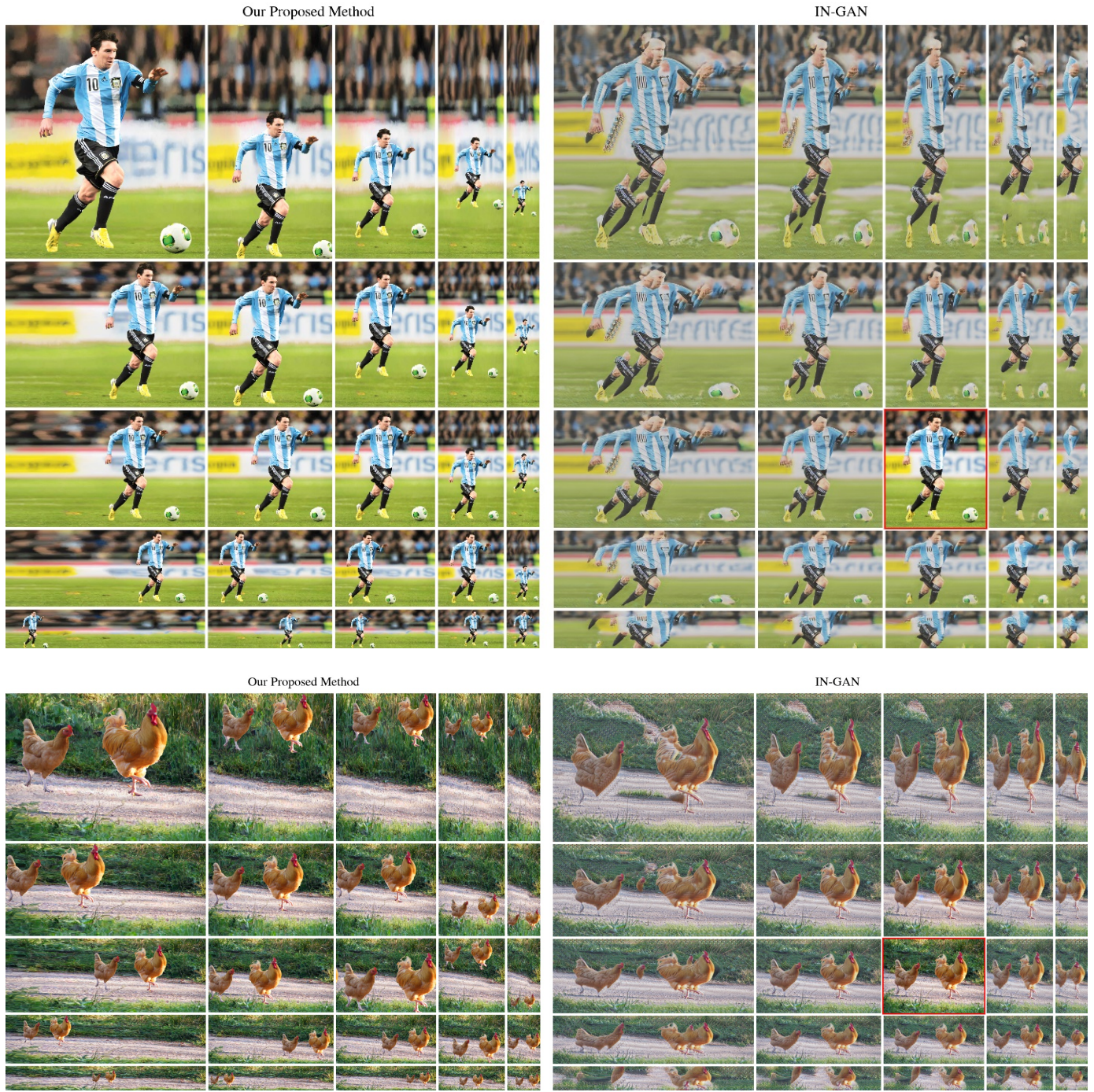

**Fig. 8.** Comparing results of our proposed method with INGAN [3] in multiple target sizes for two input images adapted from [29] in actual size. The image with the red border shows the input.

## a) Aesthetic image quality assessment

### Image 1

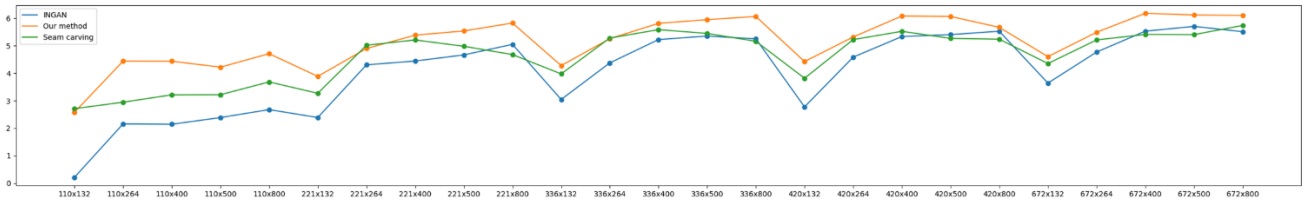

### Image 2

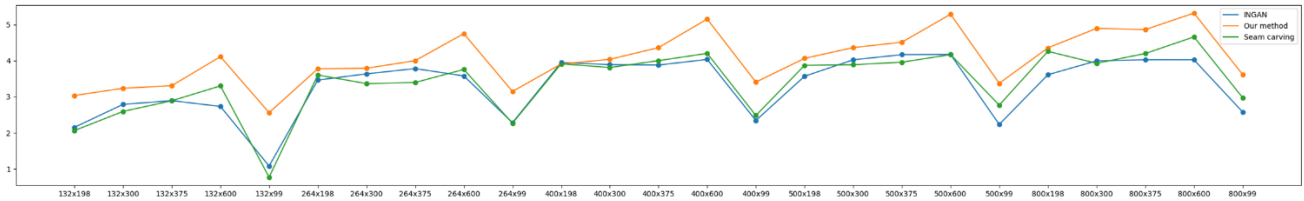

### Image 3

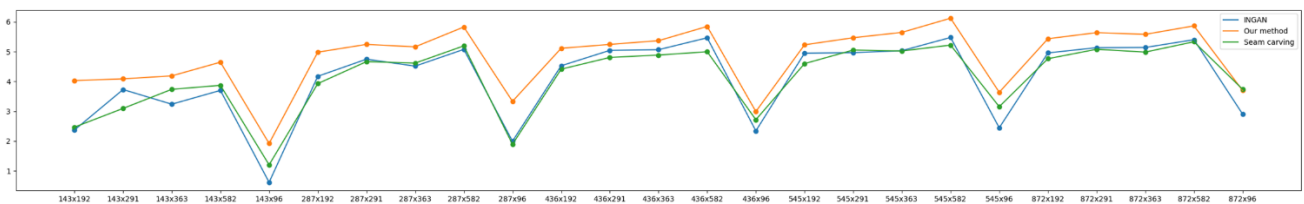

## b) Normal image quality assessment

### Image 1

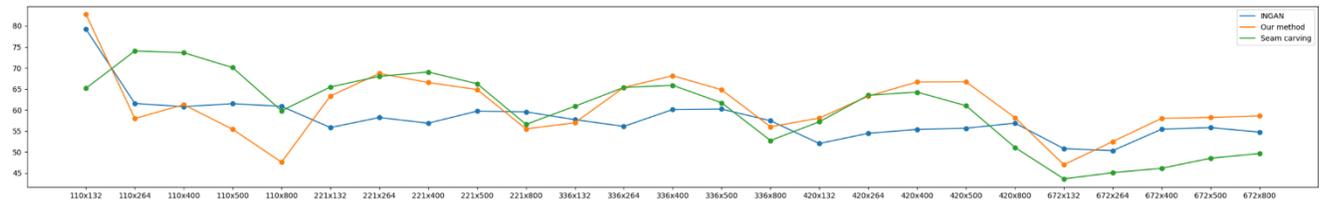

### Image 2

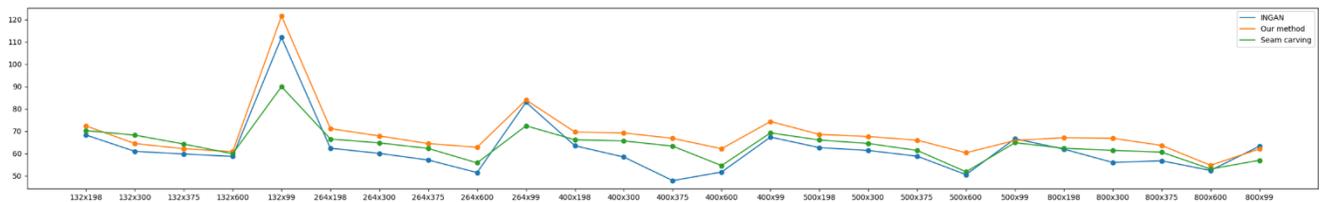

### Image 3

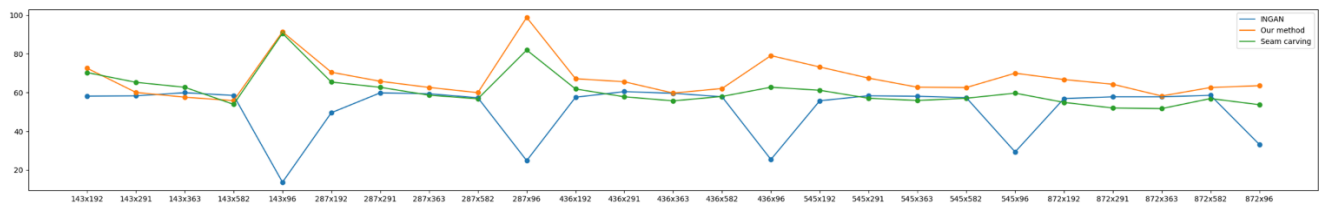

**Fig. 9.** Comparing a) aesthetic [25] and b) normal [31] scores, obtained from 25 different sizes. The horizontal axes show the retargeted image height × width, and the vertical axes show retargeted image score for each method.

# 6. Conclusion

In this work, we presented a novel content-aware image retargeting framework. Due to the absence of suitable retargeting datasets to train learning methods in this category, deep learning methods do not achieve promising results. Thus, a non-learnable technique seems more appropriate for this vision application. This is, of course, in contrast with most vision tasks.

In an image retargeting problem, multiple aspects should be considered. These concepts include object importance, the amount of distortion in the output image, and the aesthetic quality of the retargeted image. Nevertheless, a new understanding of the problem and redefining the problem boundaries could yield interesting results in this field. As we have shown by considering the factors mentioned above, our method performs well at a content-aware image retargeting problem.